\documentclass[10pt,twocolumn,letterpaper]{article}

\usepackage{iccv}
\usepackage{times}
\usepackage{epsfig}
\usepackage{graphicx}
\usepackage{amsmath}
\usepackage{amssymb}
\usepackage{booktabs}
\usepackage{multirow}
\usepackage{float}
\usepackage[table,xcdraw]{xcolor}

\usepackage[pagebackref=true,breaklinks=true,letterpaper=true,colorlinks,bookmarks=false]{hyperref}

\iccvfinalcopy %

\def\iccvPaperID{10157} %

\newcommand{\dalle}{DALL·E~$2$ }

\ificcvfinal\fi

\begin{document}

\title{Inspecting the Geographical Representativeness \\ 
of Images from Text-to-Image Models}

\author{Abhipsa Basu\\
{\tt\small abhipsabasu@iisc.ac.in}
\and
R. Venkatesh Babu\\
{\tt\small venky@iisc.ac.in}\\
Indian Institute of Science, Bangalore
\and
Danish Pruthi\\
{\tt\small danishp@iisc.ac.in}
}

\maketitle
\ificcvfinal\thispagestyle{empty}\fi

\begin{abstract}

Recent progress in generative models has 
resulted in models that produce both realistic as well as 
relevant images for most textual inputs. 
These models 
are being used to 
generate millions of images everyday,
and hold the potential to drastically impact areas 
such as generative art, digital marketing and data augmentation.
Given their outsized impact, 
it is important 
to ensure that the generated content 
reflects the artifacts and surroundings across the globe, 
rather than over-representing certain parts of the world. 
In this paper, 
we measure the geographical representativeness 
of common nouns (e.g., a house) generated through
\dalle and Stable Diffusion models
using a crowdsourced study comprising $540$ participants
across $27$ countries.
For deliberately underspecified inputs without country names,
the generated images 
most reflect the surroundings 
of the 
United States followed by India, and 
the top generations rarely reflect 
surroundings from all other countries (average score less than $3$ out of $5$).
Specifying the country names in the input
increases the representativeness by $1.44$ points on average for \dalle 
and $0.75$ for Stable Diffusion, 
however, the overall scores for many countries still remain low,
highlighting the need for 
future models to be more geographically inclusive.
Lastly, 
we examine the feasibility of quantifying the 
geographical representativeness of generated images without  
conducting user studies.

\end{abstract}

\section{Introduction}
\label{sec:intro}

Over the last year, 
the quality of 
text-to-image generation systems 
has remarkably improved~\cite{ramesh2022hierarchical, yu2022scaling, rombach2021highresolution, ruiz2022dreambooth}.
The generated images 
are more realistic 
and relevant to the textual input. 
This progress in text-to-image synthesis 
is partly fueled 
by the sheer scale of models and datasets used to train them, 
and partly by 
the architectural advancements including 
Transformers~\cite{vaswani2017attention} and Diffusion models~\cite{ho2020denoising}.
Given the impressive generation capabilities that these 
models display, 
such models have captured the interest of 
researchers and general public alike. 
For instance, DALL·E~$2$ 
is being used by over 
$1.5$ million users to generate
more than $2$ million images per day
for applications 
including art creation, 
image editing, 
digital marketing and data augmentation~\cite{openai_usage}. 

\begin{figure}[t]
    \centering
    \includegraphics[width = 0.75\columnwidth]{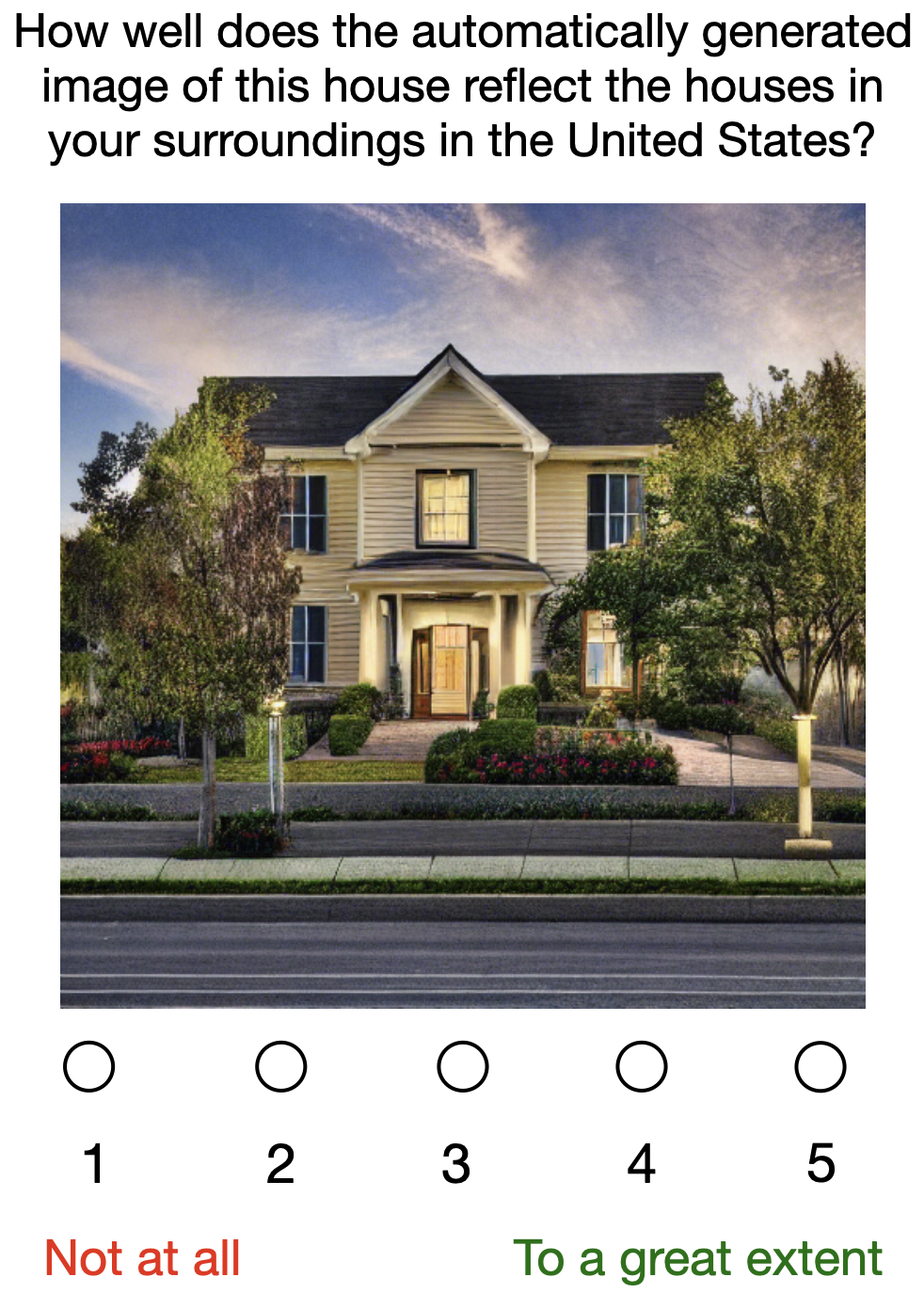}
    \caption{An illustrative question from our study, where a participant (in this case, from the United States) is presented with an image of a common noun (a house), generated from the Stable Diffusion model. The participant is asked to rate the generated image on how well it reflects the houses in their surroundings.}
    
    \label{fig:illustrative_example}
\end{figure}

Despite the broad appeal of text-to-image models,
there are looming concerns about
how these models may exhibit and amplify existing societal biases. 
These concerns stem 
from the fact that image generation 
models are trained 
on large swaths
of image-caption pairs mined from the internet,
which is known to 
be rife with toxic, stereotyping, and biased content.
Further, internet access itself is 
unequally distributed,
leading to underrepresentation and exclusion of voices from  
developing and poor nations~\cite{pew_research_internet_access, world_bank_internet_access}

There exists a wide body 
of work demonstrating 
biases in large language and vision models~\cite{hendricks2018women, wolfe2022markedness, ross2020measuring, steed2021image}, 
and some recent 
work investigates 
text-to-image models 
for biases related to
representation of race, 
gender 
and 
occupation~\cite{cho2022dall, bianchi2022easily}. 
Another important---and often overlooked---aspect 
of inclusive representation is 
\emph{geographical representation}. 
For
such
systems to be geographically representative,
they should generate images 
that represent the objects and surroundings 
of different nations in the world, 
and refrain from 
overrepresenting certain nations
and contributing to their hegemony. 
For instance, a typical house in 
the United States looks different from one in Japan.
Often the input descriptions to 
text-to-image models are underspecified,
leaving the models to fill in the missing details.
In such underspecified descriptions, there is 
an increasing risk that models 
overrepresent certain demographics~\cite{hutchinson2022underspecification}.
In addition to representational harms, 
biased image generation systems
can also cause allocational harms 
as such systems are 
used to augment  datasets, 
which run the risk of 
further propagating existing biases.
Further, the experience 
of using systems 
that underrepresent 
certain areas
would likely be unpleasant 
for the residents of those regions.

In this paper, 
we measure the  
degree to which the 
text-to-image-generation 
systems 
produce images 
that reflect the artifacts and surroundings of 
participants from different parts of the world (\S\ref{sec:approach}).
To answer this question, 
we conduct a user study 
involving $540$ participants from $27$ different countries.
We present each user 
$80$ images of common nouns generated from   
DALL·E~$2$~\cite{ramesh2022hierarchical} and Stable Diffusion~\cite{rombach2021highresolution} models.
Half of the presented images 
are generated by specifying the 
country of the participant in the input, 
and the remaining 
images are deliberately underspecified 
to examine the default generations.
The users evaluate the presented images 
based on a $5$-point Likert 
scale indicating how well do the generated 
images reflect the given entity in their physical surroundings (see Figure~\ref{fig:illustrative_example}).  
We also ask respondents 
to score generated images on (i) how realistic they look, and (ii) how
the realism impacted their scores about geographical representativeness.

Overall, we find that 
the geographical representativeness 
of images for many countries 
is considerably low (\S\ref{sec:results}). 
In the unspecified case, i.e., without any country name
in the input, 
we find that the generated images most reflect artifacts
from the United States (average geographical representativeness score of 3.35 out of 5),
followed by India (score of 3.23) and Canada (score of 2.82), 
and least reflect the nouns from Greece, Japan and New Zealand (with scores less than or around 2.0). 
Out of 27 countries, 25 countries
have a score of less than 3 for both \dalle and Stable Diffusion models.
When we specify the 
country name
in the input prompt, 
the average score over all the studied countries increases to 
$3.49$ (from $2.39$ in the unspecified case). 
However, these scores suggest that there is
room for future text-to-image models
to produce more geographically representative content.
Between DALL·E~$2$ and Stable Diffusion, 
we find DALL·E~$2$ to be better 
at generating geographically representative content 
when we specify country names, 
but
we observe 
no statistically significant 
difference 
in the underspecified case.\footnote{Note that the scope and focus of our study is solely
on measuring the extent 
of geographical representativeness 
for both country-specified and unspecified prompts,
rather than 
finding better ways to prompt the model,
or improve the model to produce more geographically 
inclusive content.}
We find that 
the participants' 
ratings about the realism of the 
images
are correlated 
with their scores about the geographical representativeness.

Finally, we examine the feasibility of automating 
the process of 
quantifying the geographical representativeness of text-to-image generation models through two different ways (\S\ref{sec:clip}).
First, we consider  
the similarity of a country-specific textual prompt and the test image using CLIP, a pre-trained text-image alignment model~\cite{radford2021learning}. 
Second,
we evaluate the viability 
of using user annotations for \dalle
as a means for estimating the geographical representativeness 
for images generated through Stable Diffusion.
We find both these approaches to be inadequate 
in accurately evaluating the geographical representativeness 
of the images, emphasizing the need for a user study.
We conclude with a discussion on  limitations of our work, and suggestions for future research in this area (\S\ref{sec:limitations}).

\section{Approach}
\label{sec:approach}

\paragraph{Geographical Representativeness.}
We 
present
crowdworkers 
from different 
countries 
with 
several model-generated images 
of common nouns, 
and for each image, 
we ask them to rate 
on a scale of $1$-$5$ about 
how well do the generated images 
reflect their surroundings. 
Geographical representativeness ($\textbf{GR}$) of the model $m$ for country $c$, is then defined as the average rating 
participants from that country provide to the model generated images of common nouns ($\mathcal{N}$), using a corresponding set of input prompts ($\mathcal{P}$). 
Similarly, we define the realism, $\text{R} (c, m, p)$, as the average of realism ratings given by participants from country $c$ to images generated by model $m$ using a prompt $p$.

\paragraph{Research Questions.} Using the above notions of geographical representativeness and realism, we ask: 
\begin{itemize}
    \item \textbf{RQ1}: Are the images generated using \dalle and Stable Diffusion geographically representative? Do they over-represent rich or populous nations?
    \item \textbf{RQ2}: To what extent does specifying the country name in the input improve the representativeness?
    \item \textbf{RQ3}: Does the realism of images impact participants' ratings about the geographical representativeness? 
    \item \textbf{RQ4}: How feasible is it to automatically assess the geographical representativeness of generated images? 
\end{itemize}
\paragraph{Selected Countries.} 
We reach out to residents of $88$ countries using 
Amazon Mechanical Turk (AMT)\footnote{\url{https://www.mturk.com}}
and 
Prolific\footnote{\url{https://www.prolific.co}}
crowdsourcing platforms. 
However, a large majority of 
crowdworkers 
belong to only a few countries, 
and 
we eventually end up with sufficient responses only from $27$
countries. 
We sample the $88$ countries 
using weighted random sampling 
where each nation was weighted by its population. 
The final set of $27$ countries (denoted by $\mathcal{C}$)
includes: 
the United States of America,
Canada, 
Mexico,
Brazil,
Chile, 
the United Kingdom, 
Italy, 
Spain, 
Greece,
Japan,
Korea,
India,
Israel,
Australia, 
South Africa,
Belgium,
Poland,
Portugal,
Germany,
France,
Latvia,
Hungary,
the Czech Republic,
Estonia,
New Zealand,
Finland,
and
Slovenia.

\paragraph{Chosen Artifacts.}
To curate a list of 
diverse but common artifacts,
we extract the most common nouns from the 
popular Conceptual Captions dataset~\cite{sharma2018conceptual} which contains image-caption pairs, 
used for training various vision+language systems~\cite{lu2019vilbert, su2019vl, li2020oscar, ramesh2021zero}. 
We use a POS-tagger from the NLTK library
to extract the nouns, and sort them by
decreasing order of their frequency. 
We choose the $10$ most common
nouns 
after manually excluding nouns that are universal in nature (e.g., sky, sun).
The final list of $10$ common nouns, denoted by $\mathcal{N}$, includes
city, beach, house, festival, road, dress, flag, park, wedding, 
and 
kitchen.

\paragraph{Input Prompts.}
As mentioned earlier, we use two types of queries for image synthesis. 
For half of the queries, 
we include the country name,
and for the remaining half, 
we do not specify any country name (to assess the default generations). 
When specifying the country name, we modify the query to 
``high definition image of a typical \texttt{[artifact]} in \texttt{[country]}'', 
where 
we include the word typical 
to generate the most common form of the 
concept in the specified country. We denote such queries by $p_c$, where $c$ refers to the country name in question.
For the underspecified case, our query is  ``high definition image of a \texttt{[artifact]}'', which we denote by $p$. 
We use the same 
prompt for both \dalle and Stable Diffusion models.

\paragraph{Questionnaire Details.} 
For each of the $10$ nouns, 
we generate $8$ images, 
$4$ using \dalle and $4$ from Stable Diffusion. 
Overall for a given country, our survey comprises $80$ images.
Participants 
are not privy to the details of the models, 
and do not know which images were generated from which model.
For each image, we ask each participant: 
``How well does the automatically generated image of this \texttt{[artifact]}
reflect the \texttt{[artifact]} in your surroundings in \texttt{[country]}?''. (See Figure~\ref{fig:illustrative_example}).
For each question, 
participants 
mark their responses using a $5$-point Likert scale, where where $1$ indicates ``not at all'', and $5$ represents ``to a great extent''. 
After the $80$ questions, 
we ask the users
to rate the photo-realism 
of the generated images on a scale of $1$-$5$, and  
how it impacted their scores about geographical representativeness.
We pay AMT participants 
based on the estimated hourly income of crowdworkers in their respective countries. %
For participants from Prolific, we pay 
them a platform-set minimum of $6.91$ USD per hour.

\begin{figure}[t]
    \centering
    \includegraphics[width = 0.85\columnwidth]{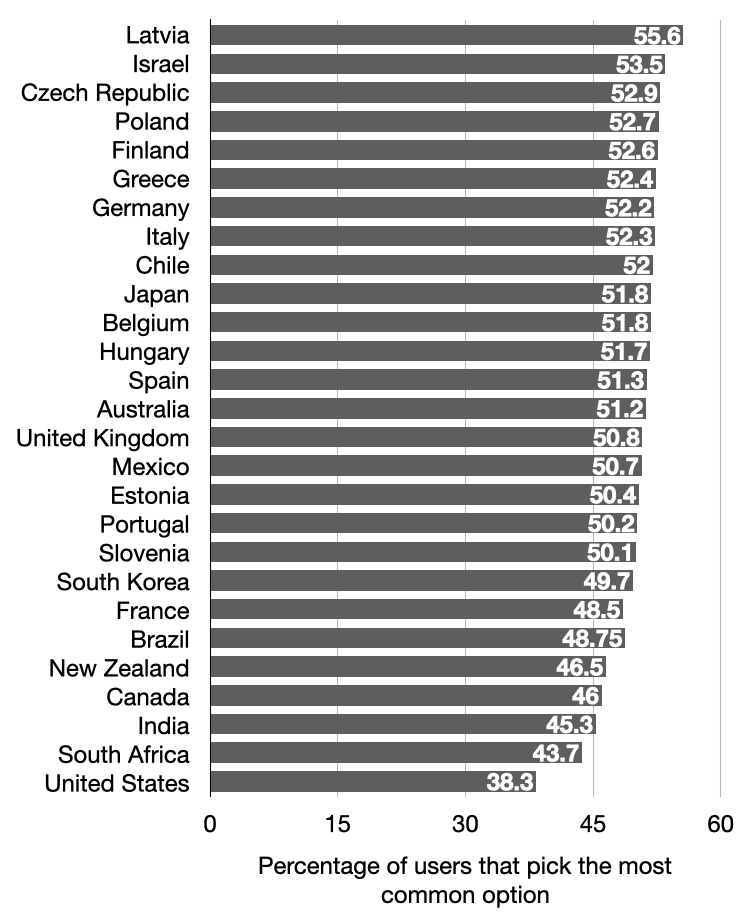}
    \caption{\textbf{Agreement among participants.} We plot the percentage of participants from each country that choose the most common option (for that country). We see that there is a considerable agreement among respondents, as about half the participants in many countries agree on one out of five options.}
    \label{fig:agreement}
\end{figure}

\paragraph{Validating Responses.} 
To verify if the participants answered the questions earnestly, we include $4$
trick questions which are presented in the same format. Two of these 
trick questions 
inquire about apples and milk, whereas 
the corresponding images are of mangoes and water. 
Therefore, we expect participants 
to mark a low score for these two questions. 
For the other two trick questions, 
we ask about a pen and sun, and include
images of the same, and expect the users to mark a high score. 
We discard the responses from participants 
who do not pass these checks. 
While the crowdsourcing platforms 
allow us to target users from a given country,
we re-confirm with participants if they indeed 
reside (or have lived) in the specified countries.

\paragraph{Inter-rater Agreement.}
We compute (for each country) 
the percentage of participants 
who opted for the most selected option.
We observe a high agreement among participants; 
for $19$ out of the $27$ studied countries
we see that the most common option is picked by over $50\%$ of the respondents (Figure~\ref{fig:agreement}).
The agreement would be (on an average) $20\%$ if 
participants marked options arbitrarily.
The percentages in Figure~\ref{fig:agreement} 
demonstrate some degree of consensus among participants.
Further, we observe the highest agreement 
for images of flags ($81\%$) and the least agreement for kitchens ($41\%$).

\section{Results}
\label{sec:results}

\begin{table*}[ht]
\centering
\caption{\textbf{Geographic Representativeness}. We tabulate the geographical representativeness scores for \dalle (D2), Stable Diffusion (SD) and combination of both the models (Overall) for different countries, both for the case when the model is prompted using the country name and without it. In the unspecified case, we observe that the scores is highest for United States, followed by India (scores greater than $3.2$ out of $5$), but low for many other countries. We observe a consistent improvement in the scores when we include the country names.}
\label{tab:all-together}
\begin{tabular}{@{}lclcccc@{}}
\toprule
\multirow{3}{*}{Countries} & \multicolumn{2}{c}{Overall}                                                                                                   & \multicolumn{2}{c}{\dalle}                                                        & \multicolumn{2}{c}{Stable Diffusion}                                            \\ \cmidrule(l){2-7} 
                           & w/ country                                                   & \multicolumn{1}{c}{Unspecified}                                & w/ country                            & Unspecified                               & w/ country                            & Unspecified                             \\
                           & \multicolumn{1}{l}{\textbf{$\text{GR}(c, \cdot, \cdot, p)$}} & \multicolumn{1}{c}{\textbf{$\text{GR}(c, \cdot, \cdot, p_c)$}} & $\textbf{GR}(c, \text{D2}, \cdot, p)$ & {$\textbf{GR}(c, \text{D2}, \cdot, p_c)$} & $\textbf{GR}(c, \text{SD}, \cdot, p)$ & $\textbf{GR}(c, \text{SD}, \cdot, p_c)$ \\ \midrule
US                         & $3.54$     \footnotesize{$\pm 0.23$}                           & $3.35$ \footnotesize{$\pm 0.18$}                                 & $3.56$ \footnotesize{$\pm 0.29$}        & $3.24$   \footnotesize{$\pm 0.27$}          & $3.51$  \footnotesize{$\pm 0.25$}       & $3.46$ \footnotesize{$\pm 0.27$}          \\
India                      & $3.74$     \footnotesize{$\pm 0.26$}                           & $3.24$     \footnotesize{$\pm 0.41$}                             & $4.00$ \footnotesize{$\pm 0.22$}        & $3.44$     \footnotesize{$\pm 0.49$}        & $3.48$   \footnotesize{$\pm 0.49$}      & $3.03$ \footnotesize{$\pm 0.41$}          \\
Canada                     & $3.62$      \footnotesize{$\pm 0.40$}                          & $2.82$  \footnotesize{$\pm 0.51$}                                & $3.78$ \footnotesize{$\pm 0.55$}        & $2.73$      \footnotesize{$\pm 0.59$}       & $3.47$  \footnotesize{$\pm 0.52$}       & $2.91$ \footnotesize{$\pm 0.66$}          \\
South Africa               & $3.25$      \footnotesize{$\pm 0.30$}                          & $2.74$       \footnotesize{$\pm 0.40$}                           & $3.49$ \footnotesize{$\pm 0.57$}        & $2.70$    \footnotesize{$\pm 0.44$}         & $3.02$   \footnotesize{$\pm 0.52$}      & $2.78$ \footnotesize{$\pm 0.58$}          \\
Brazil                     & $3.70$      \footnotesize{$\pm 0.26$}                          & $2.69$ \footnotesize{$\pm 0.55$}                                 & $4.00$  \footnotesize{$\pm 0.38$}       & $2.65$  \footnotesize{$\pm 0.78$}           & $3.40$  \footnotesize{$\pm 0.23$}       & $2.72$ \footnotesize{$\pm 0.56$}          \\
UK                         & $3.82$    \footnotesize{$\pm 0.38$}                            & $2.65$ \footnotesize{$\pm 0.49$}                                 & $4.14$ \footnotesize{$\pm 0.53$}        & $2.41$       \footnotesize{$\pm 0.61$}      & $3.48$  \footnotesize{$\pm 0.56$}       & $2.88$ \footnotesize{$\pm 0.80$}          \\
Mexico                     & $3.83$        \footnotesize{$\pm 0.26$}                        & $2.59$  \footnotesize{$\pm 0.56$}                                & $4.18$ \footnotesize{$\pm 0.30$}        & $2.74$       \footnotesize{$\pm 0.72$}      & $3.49$   \footnotesize{$\pm 0.57$}      & $2.45$  \footnotesize{$\pm 0.64$}         \\
Spain                      & $3.44$     \footnotesize{$\pm 0.29$}                           & $2.46$ \footnotesize{$\pm 0.44$}                                 & $3.62$ \footnotesize{$\pm 0.38$}        & $2.29$        \footnotesize{$\pm 0.65$}     & $3.26$   \footnotesize{$\pm 0.53$}      & $2.63$ \footnotesize{$\pm 0.66$}          \\
Portugal                   & $3.73$    \footnotesize{$\pm 0.29$}                            & $2.46$ \footnotesize{$\pm 0.47$}                                 & $4.02$ \footnotesize{$\pm 0.40$}        & $2.47$      \footnotesize{$\pm 0.73$}       & $3.44$   \footnotesize{$\pm 0.61$}      & $2.45$ \footnotesize{$\pm 0.54$}          \\
Italy                      & $3.58$    \footnotesize{$\pm 0.47$}                            & $2.40$ \footnotesize{$\pm 0.49$}                                 & $3.66$ \footnotesize{$\pm 0.66$}        & $2.40$      \footnotesize{$\pm 0.70$}       & $3.50$   \footnotesize{$\pm 0.66$}      & $2.39$ \footnotesize{$\pm 0.63$}          \\
Belgium                    & $3.49$    \footnotesize{$\pm 0.43$}                            & $2.40$ \footnotesize{$\pm 0.52$}                                 & $3.76$ \footnotesize{$\pm 0.71$}        & $2.28$      \footnotesize{$\pm 0.57$}       & $3.21$   \footnotesize{$\pm 0.61$}      & $2.52$ \footnotesize{$\pm 0.80$}          \\
France                     & $3.32$    \footnotesize{$\pm 0.34$}                            & $2.34$ \footnotesize{$\pm 0.44$}                                 & $3.54$ \footnotesize{$\pm 0.67$}        & $2.38$      \footnotesize{$\pm 0.70$}       & $3.09$   \footnotesize{$\pm 0.47$}      & $2.30$ \footnotesize{$\pm 0.52$}          \\
Poland                     & $3.62$    \footnotesize{$\pm 0.30$}                            & $2.29$ \footnotesize{$\pm 0.44$}                                 & $4.14$ \footnotesize{$\pm 0.39$}        & $2.23$      \footnotesize{$\pm 0.59$}       & $3.10$   \footnotesize{$\pm 0.66$}      & $2.35$ \footnotesize{$\pm 0.70$}          \\
Germany                    & $3.64$    \footnotesize{$\pm 0.35$}                            & $2.26$ \footnotesize{$\pm 0.45$}                                 & $4.03$ \footnotesize{$\pm 0.30$}        & $2.04$      \footnotesize{$\pm 0.46$}       & $3.26$   \footnotesize{$\pm 0.70$}      & $2.49$ \footnotesize{$\pm 0.78$}          \\
Australia                  & $3.35$   \footnotesize{$\pm 0.45$}                             & $2.26$    \footnotesize{$\pm 0.46$}                              & $3.55$ \footnotesize{$\pm 0.74$}        & $2.10$    \footnotesize{$\pm 0.49$}         & $3.15$  \footnotesize{$\pm 0.66$}       & $2.41$ \footnotesize{$\pm 0.65$}          \\
Czech Republic             & $3.43$    \footnotesize{$\pm 0.48$}                            & $2.25$ \footnotesize{$\pm 0.52$}                                 & $3.68$ \footnotesize{$\pm 0.50$}        & $2.18$      \footnotesize{$\pm 0.64$}       & $3.18$   \footnotesize{$\pm 0.79$}      & $2.31$ \footnotesize{$\pm 0.66$}          \\
Hungary                    & $3.41$    \footnotesize{$\pm 0.49$}                            & $2.24$ \footnotesize{$\pm 0.55$}                                 & $3.65$ \footnotesize{$\pm 0.59$}        & $2.06$      \footnotesize{$\pm 0.52$}       & $3.18$   \footnotesize{$\pm 0.74$}      & $2.42$ \footnotesize{$\pm 0.76$}          \\
New Zealand                & $3.10$    \footnotesize{$\pm 0.44$}                            & $2.23$ \footnotesize{$\pm 0.36$}                                 & $3.10$ \footnotesize{$\pm 0.76$}        & $2.24$      \footnotesize{$\pm 0.70$}       & $3.11$   \footnotesize{$\pm 0.49$}      & $2.22$ \footnotesize{$\pm 0.39$}          \\
Estonia                    & $3.36$    \footnotesize{$\pm 0.25$}                            & $2.22$ \footnotesize{$\pm 0.33$}                                 & $3.89$ \footnotesize{$\pm 0.58$}        & $2.18$      \footnotesize{$\pm 0.51$}       & $2.84$   \footnotesize{$\pm 0.49$}      & $2.26$ \footnotesize{$\pm 0.49$}          \\
Slovenia                   & $3.29$    \footnotesize{$\pm 0.46$}                            & $2.21$ \footnotesize{$\pm 0.43$}                                 & $3.48$ \footnotesize{$\pm 0.49$}        & $2.19$      \footnotesize{$\pm 0.45$}       & $3.10$   \footnotesize{$\pm 0.69$}      & $2.23$ \footnotesize{$\pm 0.65$}          \\
Chile                      & $3.12$   \footnotesize{$\pm 0.40$}                             & $2.15$  \footnotesize{$\pm 0.42$}                                & $3.62$ \footnotesize{$\pm 0.64$}        & $2.26$       \footnotesize{$\pm 0.69$}      & $2.62$   \footnotesize{$\pm 0.57$}      & $2.04$  \footnotesize{$\pm 0.58$}         \\
Israel                     & $3.14$     \footnotesize{$\pm 0.39$}                           & $2.15$  \footnotesize{$\pm 0.49$}                                & $3.62$ \footnotesize{$\pm 0.67$}        & $2.10$     \footnotesize{$\pm 0.67$}        & $2.66$  \footnotesize{$\pm 0.59$}       & $2.19$ \footnotesize{$\pm 0.64$}          \\
South Korea                & $3.49$     \footnotesize{$\pm 0.24$}                           & $2.10$    \footnotesize{$\pm 0.39$}                              & $3.92$ \footnotesize{$\pm 0.45$}        & $2.24$         \footnotesize{$\pm 0.66$}    & $3.06$  \footnotesize{$\pm 0.47$}       & $1.96$ \footnotesize{$\pm 0.44$}          \\
Latvia                     & $3.52$    \footnotesize{$\pm 0.34$}                            & $2.10$ \footnotesize{$\pm 0.49$}                                 & $4.11$ \footnotesize{$\pm 0.44$}        & $1.87$      \footnotesize{$\pm 0.52$}       & $2.93$   \footnotesize{$\pm 0.46$}      & $2.32$ \footnotesize{$\pm 0.67$}          \\
Finland                    & $3.62$    \footnotesize{$\pm 0.30$}                            & $2.03$ \footnotesize{$\pm 0.34$}                                 & $3.93$ \footnotesize{$\pm 0.54$}        & $1.95$      \footnotesize{$\pm 0.44$}       & $3.30$   \footnotesize{$\pm 0.59$}      & $2.10$ \footnotesize{$\pm 0.48$}          \\
Japan                      & $3.55$     \footnotesize{$\pm 0.32$}                           & $1.95$  \footnotesize{$\pm 0.40$}                                & $3.97$ \footnotesize{$\pm 0.37$}        & $1.97$  \footnotesize{$\pm 0.49$}           & $3.13$  \footnotesize{$\pm 0.59$}       & $1.94$ \footnotesize{$\pm 0.48$}          \\
Greece                     & $3.43$   \footnotesize{$\pm 0.46$}                             & $1.94$   \footnotesize{$\pm 0.49$}                               & $3.65$ \footnotesize{$\pm 0.48$}        & $1.92$       \footnotesize{$\pm 0.66$}      & $3.22$   \footnotesize{$\pm 0.69$}      & $1.97$ \footnotesize{$\pm 0.48$}          \\ \hline
Average                    & $3.49$ \footnotesize{$\pm 0.06$}                               & $2.39$  \footnotesize{$\pm 0.08$}                                & $3.78$ \footnotesize{$\pm 0.09$}        & $2.34$    \footnotesize{$\pm 0.10$}         & $3.19$   \footnotesize{$\pm 0.10$}      & $2.44$     \footnotesize{$\pm 0.10$}      \\ \bottomrule
\end{tabular}
\end{table*}

In this section, we share the findings of our study. First, we discuss the metrics of interest, and then answer the four research questions posed in Section~\ref{sec:approach}.

\subsection{Metrics}
Below, we define a few notations that we use for evaluating the user ratings. 
Remember from Section~\ref{sec:approach} that we defined $\textbf{GR} (c, m, n, p)$ as the 
geographical representativeness score assigned by participants from country $c$ to images generated for noun $n$ from model $m$ using prompt $p$.

\begin{figure*}[ht]
    \centering
    \includegraphics[width = 0.85 \textwidth]{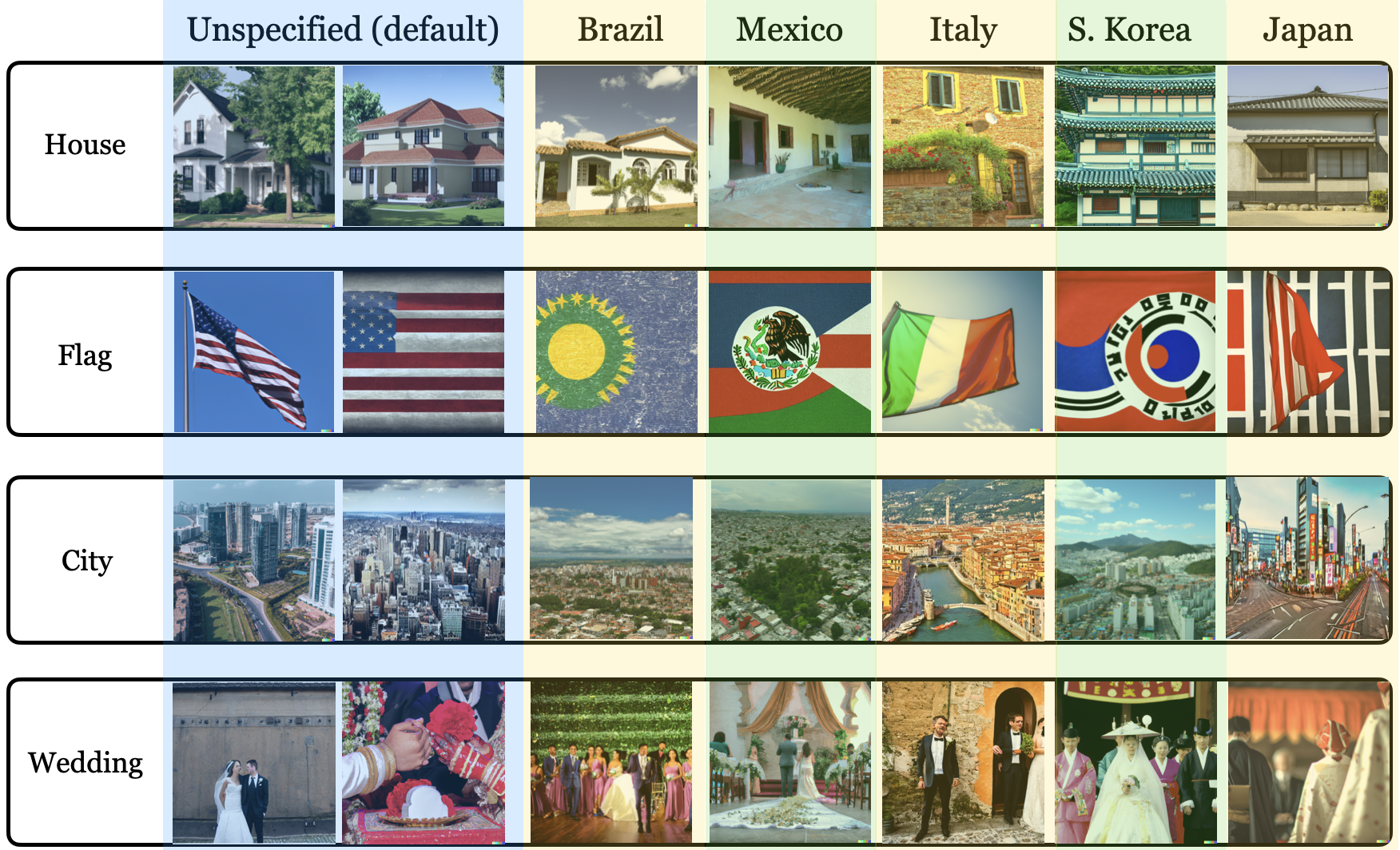}
    \caption{Qualitative examples of images of four common nouns generated by \dalle and Stable Diffusion models. Through these examples and others, we see that the default generations often reflect artifacts from US and Canada. For example, the average score (in unspecified case) for the images of houses generated through \dalle is $3.95$  for US and Canada, and $2.09$ for the remaining countries.}
    
    \label{fig: more_examples}
\end{figure*}

\begin{itemize}
\setlength\itemsep{0.05mm}
    \item  $\textbf{GR} (c, m, \cdot, p_c )$: Average ratings that participants of a country $c$ assign for geographical representativeness of images generated by model $m$ across all nouns in $\mathcal{N}$. Here, we use a country-specific prompt ($p_c$).
    \item  $\textbf{GR} (c, m, \cdot, p)$: Average ratings that participants of a country $c$ assign for geographical representativeness of images generated by model $m$ across all nouns. The prompt $p$ \textbf{does not} specify the country name.%
    \item $\textbf{GR} (\cdot, m, n, p_c )$: Average ratings that participants  for all countries in $\mathcal{C}$ assign for geographical representativeness (GR) of images of noun $n$, generated by model $m$. Here, we use a country-specific prompt ($p_c$).  
    \item $\textbf{GR} (\cdot, m, n, p)$: Average ratings that participants from all countries in $\mathcal{C}$ assign for geographical representativeness of images of noun $n$, generated by model $m$. The prompt $p$ \textbf{does not} specify the country name. %
\end{itemize}

Analogously, we define $\textbf{R} (c, m, n, p_c)$ and $\textbf{R} (c, m, n, p)$ as the average realism score for generated images using country specific ($p_c$) and unspecific prompt ($p$) respectively.

\subsection{Geographical Representativeness}

Here, we elaborate on the extent to which 
the generated artifacts are geographically representative (\textbf{RQ1} in Section \ref{sec:approach}). 
We compute the geographical representativeness scores for each country, averaged over the two models for the images generated by prompts that do not specify the country name, i.e., \textbf{$\text{GR}(c, \cdot, \cdot, p)$}.
We present these results in Table~\ref{tab:all-together}.
From the table, 
we can see that out of the $27$ countries, 
$25$ have a score 
lower than $3$ 
(on a scale of $1$ to $5$), 
indicating that participants from 
most of the studied countries 
do not feel that the 
generated images 
reflect their surroundings to a large extent. 
The only countries to obtain 
scores higher than $3$
are the United States ($3.35$)
and India ($3.23$). Interestingly, for DALL·E~$2$, India obtains the highest score ($3.44$) followed by the United States ($3.24$).
The overall least scores are assigned by 
participants from 
Greece ($1.94$), 
Japan ($1.95$) and Finland ($2.03$). 
The average score across the studied $27$ countries is $2.39$.

To answer the follow up questions posed in the \textbf{RQ1}, about whether 
the artifacts generated are
more representative of richer and populous nations: 
\begin{enumerate}
\setlength\itemsep{0.05mm}
    \item We find no correlation between the degree of geographical representativeness of the generated images for the studied countries and their per-capita GDP. The Pearson correlation coefficient, $\rho$, is $-0.03$. Moreover, after separating the country pool into the ``Rich West'' countries\footnote{As defined per: \url{https://worldpopulationreview.com/country-rankings/western-countries}} and others, we evaluate if average GR scores of the two groups are different, but we find no statistically significant difference. We acknowledge and speculate that we may observe different trends if the study included participants from many other developing countries. However, significantly improving the coverage of the study is challenging (see Section~\ref{sec:limitations}).  
    \item We observe that the geographical representativeness scores of the $27$ countries is positively correlated with their population ($\rho = 0.64$). This may suggest that the datasets used to pre-train the chosen models contain many images from residents of populous countries. 
    
\end{enumerate}

\subsection{Effect of Country-specific Prompts}

In this subsection, we analyse the geographical representativeness of images generated by including the country name (\textbf{RQ2} in Section~\ref{sec:approach}).  
From Table~\ref{tab:all-together}, we observe that for each nation, 
mentioning its name in the prompt increases the average \textbf{GR} score for that country 
as compared to the under-specified case. 
We conduct a paired sample t-test to confirm this, 
and find that indeed there is a statistically significant increase with p-value $< 0.05$. 
Specifically, adding the country name in the textual query increases the average geographical representativeness score by over $1.44$ points for \dalle and $0.75$ for Stable Diffusion. 
Overall, for $14$ out of $27$  countries
(despite the increase upon including country names),
the geographical representation scores were between $3$ to $3.5$,
indicating a considerable headroom for future models to generate more representative artifacts.

We show illustrative examples of images generated by the unspecified and country-specific prompts in Figure~\ref{fig: more_examples}. Specifically, we show images for $5$ countries: Brazil, Mexico, Italy, Japan and South Korea, and $4$ nouns: house, city, flag and wedding. 
For each of the nouns, we show images generated by the under-specified prompts first, followed by the ones generated through country specific prompts. In the appendix, we show images generated separately by both DALL·E~$2$ \cite{ramesh2022hierarchical} and Stable Diffusion \cite{rombach2021highresolution} for all the $10$ nouns, whereas we choose one country from each continent: US, Chile, UK, Japan, South Africa, and Australia. The generated images are presented in Fig. \ref{fig:d2-1} and \ref{fig:d2-2} for DALL·E~$2$, and Fig. \ref{fig:SD-1} and \ref{fig:SD-2} for Stable Diffusion.

\subsection{Photo-realism of Generated Images}

We seek to answer if, and to what degree, does the photo-realism of images impact  participants' perceptions of geographical representativeness of a given artifact (\textbf{RQ3} in Section~\ref{sec:approach}). We believe that there may be an effect, as unrealistic-looking images might be perceived less geographically appropriate (in the extreme case, unrealistic-looking photos might be hard to even interpret). 
To answer this question, we ask participants to rate the realism of images generated by DALL·E~$2$ and Stable Diffusion respectively (for both the under-specified and country-specific prompts) on a Likert-scale of $1$ to $5$. Additionally, 
in the exit survey, we ask participants to self assess the impact that the realism
of images had on the scores they assigned for geographical representativeness of images.

First, we find that geographical representativeness 
and realism scores are correlated, with a Pearson correlation of 
$0.62$ for Stable Diffusion (unspecified case), and $0.47$ for the case with country names. For \dalle the correlation is not as large ($0.21$ and $0.57$ for unspecified and country-specific prompts respectively).
This is also concordant with the self-evaluation 
provided by participants, where we note 
that participants, on average, 
indicate that the realism influenced their ratings on geographical representativeness 
to a moderate extent (average score of $3.5$ on a scale of $1$-$5$).
Interestingly, we find that
that the average realism score
assigned by participants is lower (averaged over all countries) 
when the prompt excludes the country name (this difference is statistically significant  with p value $< 0.05$). Albeit, we do see that for some countries, e.g., the United States and Brazil, 
the realism scores decreases upon including the country names in the prompt. 
More details and country-wise statistics on the realism values (for all $27$ countries) can be found in 
Table \ref{tab:photorealism}. 
\begin{table*}[ht]
\centering
\caption{\textbf{Country wise photo-realism scores.} We present how the realism scores of images generated from \dalle (D2) and Stable Diffusion (SD) improve when the country name is specified in the text prompt. Additionally, in the last column we include the scores that users assign when asked about how the realism of images influenced their ratings about geographical representativeness.}
\label{tab:photorealism}
\begin{tabular}{@{}lrlrrr@{}}
\toprule
\multirow{2}{*}{Countries} & \multicolumn{2}{c}{\dalle}                                                  & \multicolumn{2}{c}{Stable Diffusion}                                    & \multirow{2}{*}{Self-Assessed}        \\ \cmidrule(lr){2-5}
                           & \multicolumn{1}{c}{w/ country}       & \multicolumn{1}{c}{Unspecified}      & \multicolumn{1}{c}{w/ country}    & \multicolumn{1}{c}{Unspecified}     &                                       \\ \midrule
United States              & 3.82     \footnotesize{$\pm 0.37$}   & 4.00  \footnotesize{$\pm 0.36$}      & 3.88    \footnotesize{$\pm 0.40$} & 4.12  \footnotesize{$\pm 0.32$}     & 3.88        \footnotesize{$\pm 0.46$} \\
Canada                     & 4.06     \footnotesize{$\pm 0.27$}   & 3.88     \footnotesize{$\pm 0.49$}   & 3.06 \footnotesize{$\pm 0.44$}    & 3.19    \footnotesize{$\pm 0.40$}   & 3.31    \footnotesize{$\pm 0.68$}     \\
Mexico                     & 4.23  \footnotesize{$\pm 0.72$}      & 3.31    \footnotesize{$\pm 0.58$}    & 3.23 \footnotesize{$\pm 0.63$}    & 2.54   \footnotesize{$\pm 0.66$}    & 3.62 \footnotesize{$\pm 0.77$}        \\
Brazil                     & 4.07    \footnotesize{$\pm 0.34$}    & 4.27  \footnotesize{$\pm 0.39$}      & 3.40  \footnotesize{$\pm 0.45$}   & 3.27    \footnotesize{$\pm 0.34$}   & 3.53   \footnotesize{$\pm 0.58$}      \\
Chile                      & 4.37  \footnotesize{$\pm 0.33$}      & 4.16 \footnotesize{$\pm 0.47$}       & 3.05 \footnotesize{$\pm 0.45$}    & 2.26   \footnotesize{$\pm 0.54$}    & 3.58      \footnotesize{$\pm 0.45$}   \\
United Kingdom             & 4.46     \footnotesize{$\pm 0.34$}   & 3.92  \footnotesize{$\pm 0.58$}      & 3.38 \footnotesize{$\pm 0.55$}    & 3.54    \footnotesize{$\pm 0.51$}   & 3.77     \footnotesize{$\pm 0.65$}    \\
Italy                      & 3.73       \footnotesize{$\pm 0.50$} & 3.20   \footnotesize{$\pm 0.62$}     & 3.47 \footnotesize{$\pm 0.52$}    & 2.60  \footnotesize{$\pm 0.58$}     & 3.40       \footnotesize{$\pm 0.69$}  \\
Spain                      & 4.11   \footnotesize{$\pm 0.37$}     & 4.00       \footnotesize{$\pm 0.75$} & 3.67 \footnotesize{$\pm 0.53$}    & 3.11  \footnotesize{$\pm 0.65$}     & 2.89   \footnotesize{$\pm 0.72$}      \\
Greece                     & 4.32 \footnotesize{$\pm 0.29$}       & 3.74    \footnotesize{$\pm 0.48$}    & 4.00   \footnotesize{$\pm 0.39$}  & 3.47   \footnotesize{$\pm 0.55$}    & 3.42  \footnotesize{$\pm 0.51$}       \\
Poland                     & 4.73 \footnotesize{$\pm 0.22$}       & 4.13    \footnotesize{$\pm 0.41$}    & 3.93   \footnotesize{$\pm 0.34$}  & 2.8   \footnotesize{$\pm 0.55$}     & 3.27  \footnotesize{$\pm 0.68$}       \\
Portugal                   & 4.35 \footnotesize{$\pm 0.40$}       & 3.88    \footnotesize{$\pm 0.56$}    & 4.11   \footnotesize{$\pm 0.46$}  & 3.29   \footnotesize{$\pm 0.51$}    & 3.47  \footnotesize{$\pm 0.77$}       \\
Belgium                    & 3.75 \footnotesize{$\pm 0.39$}       & 3.75    \footnotesize{$\pm 0.44$}    & 3.40   \footnotesize{$\pm 0.32$}  & 2.80   \footnotesize{$\pm 0.45$}    & 3.05  \footnotesize{$\pm 0.47$}       \\
Czech Republic             & 3.67 \footnotesize{$\pm 0.38$}       & 3.44    \footnotesize{$\pm 0.56$}    & 3.11   \footnotesize{$\pm 0.43$}  & 2.72   \footnotesize{$\pm 0.48$}    & 3.56  \footnotesize{$\pm 0.52$}       \\
Hungary                    & 4.26 \footnotesize{$\pm 0.35$}       & 3.84    \footnotesize{$\pm 0.47$}    & 4.00   \footnotesize{$\pm 0.36$}  & 3.53   \footnotesize{$\pm 0.42$}    & 3.21  \footnotesize{$\pm 0.65$}       \\
Slovenia                   & 3.78 \footnotesize{$\pm 0.29$}       & 3.61    \footnotesize{$\pm 0.51$}    & 3.27   \footnotesize{$\pm 0.37$}  & 2.61   \footnotesize{$\pm 0.47$}    & 3.33  \footnotesize{$\pm 0.56$}       \\
Germany                    & 4.17 \footnotesize{$\pm 0.28$}       & 3.83    \footnotesize{$\pm 0.38$}    & 3.50   \footnotesize{$\pm 0.38$}  & 3.33   \footnotesize{$\pm 0.44$}    & 4.11  \footnotesize{$\pm 0.40$}       \\
Latvia                     & 4.56 \footnotesize{$\pm 0.28$}       & 3.33    \footnotesize{$\pm 0.58$}    & 2.72  \footnotesize{$\pm 0.40$}   & 2.39   \footnotesize{$\pm 0.47$}    & 3.67  \footnotesize{$\pm 0.58$}       \\
Estonia                    & 4.21 \footnotesize{$\pm 0.31$}       & 3.47    \footnotesize{$\pm 0.42$}    & 3.00   \footnotesize{$\pm 0.36$}  & 3.11   \footnotesize{$\pm 0.44$}    & 3.74  \footnotesize{$\pm 0.50$}       \\
Finland                    & 4.16 \footnotesize{$\pm 0.39$}       & 3.95    \footnotesize{$\pm 0.42$}    & 3.26   \footnotesize{$\pm 0.46$}  & 3.00   \footnotesize{$\pm 0.51$}    & 3.74  \footnotesize{$\pm 0.60$}       \\
France                     & 4.00 \footnotesize{$\pm 0.39$}       & 3.63    \footnotesize{$\pm 0.47$}    & 3.31   \footnotesize{$\pm 0.39$}  & 3.37   \footnotesize{$\pm 0.44$}    & 3.00  \footnotesize{$\pm 0.55$}       \\
India                      & 4.31 \footnotesize{$\pm 0.26$}       & 3.92 \footnotesize{$\pm 0.47$}       & 3.62 \footnotesize{$\pm 0.47$}    & 3.85 \footnotesize{$\pm 0.37$}      & 4.16 \footnotesize{$\pm 0.44$}        \\
Japan                      & 3.89  \footnotesize{$\pm 0.37$}      & 3.67  \footnotesize{$\pm 0.46$}      & 2.94  \footnotesize{$\pm 0.52$}   & 2.78   \footnotesize{$\pm 0.55$}    & 3.22    \footnotesize{$\pm 0.50$}     \\
South Korea                & 4.32   \footnotesize{$\pm 0.29$}     & 3.42  \footnotesize{$\pm 0.47$}      & 3.21  \footnotesize{$\pm 0.47$}   & 3.05    \footnotesize{$\pm 0.61$}   & 3.89     \footnotesize{$\pm 0.38$}    \\
Israel                     & 4.54  \footnotesize{$\pm 0.34$}      & 4.15 \footnotesize{$\pm 0.36$}       & 2.92  \footnotesize{$\pm 0.58$}   & 3.31    \footnotesize{$\pm 0.69$}   & 2.62        \footnotesize{$\pm 0.75$} \\
Australia                  & 3.93 \footnotesize{$\pm 0.43$}       & 3.80    \footnotesize{$\pm 0.62$}    & 3.93  \footnotesize{$\pm 0.54$}   & 3.47 \footnotesize{$\pm 0.52$}      & 3.67       \footnotesize{$\pm 0.71$}  \\
New Zealand                & 3.75 \footnotesize{$\pm 0.44$}       & 3.40    \footnotesize{$\pm 0.56$}    & 2.85   \footnotesize{$\pm 0.38$}  & 2.65   \footnotesize{$\pm 0.49$}    & 3.35  \footnotesize{$\pm 0.52$}       \\
South Africa               & 4.06  \footnotesize{$\pm 0.61$}      & 3.94     \footnotesize{$\pm 0.50$}   & 3.06 \footnotesize{$\pm 0.66$}    & 3.44  \footnotesize{$\pm 0.55$}     & 3.50  \footnotesize{$\pm 0.65$}       \\ \midrule
Average                    & 4.14  \footnotesize{$\pm 0.08$}      & 3.73  \footnotesize{$\pm 0.10$}      & 3.38  \footnotesize{$\pm 0.10$}   & 3.09      \footnotesize{$\pm 0.11$} & 3.50     \footnotesize{$\pm 0.12$}    \\ \bottomrule
\end{tabular}
\end{table*}

\subsection{Comparison of \dalle and Stable Diffusion}
\begin{figure}[]
    \centering
    \includegraphics[width = 0.7 \columnwidth]{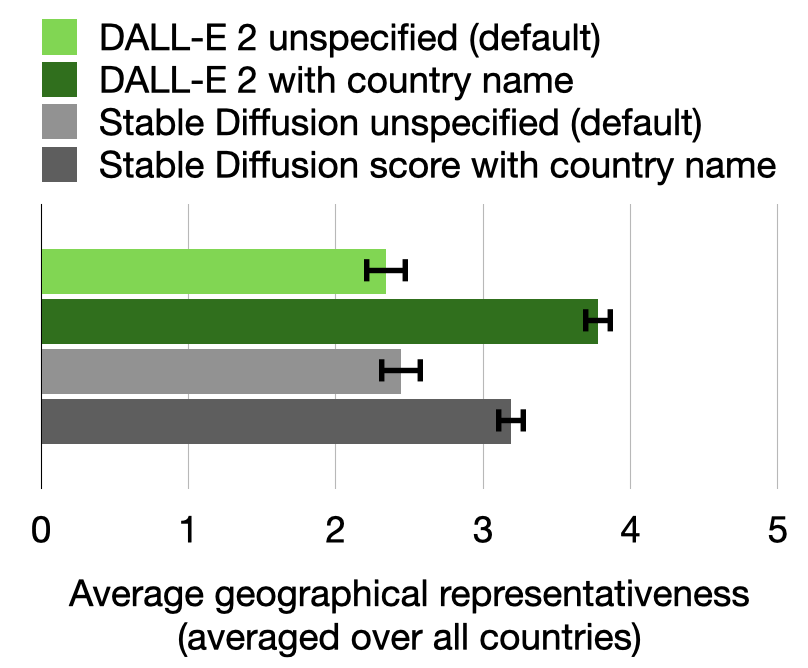}
    \caption{\textbf{DALL-E 2 vs Stable Diffusion:} Average geographical representativeness scores for images generated by \dalle and Stable Diffusion, with and without country-specific prompts.} 
    
    \label{fig:d_vs_sd}
\end{figure}

We compare \dalle vs Stable Diffusion models to see which model produces more geographically representative images (Figure~\ref{fig:d_vs_sd}). We find that
(i) for country-specific prompts, the geographical representativeness 
of images generated through \dalle are higher than those from Stable Diffusion by about $0.6$ points (and this difference is statistically significant as per a paired t-test with a p-value $<0.05$); and (ii) for country agnostic prompts, the differences are not statistically significant (see Figure~\ref{fig:d_vs_sd}). 

\section{Feasibility of Automating the Evaluation}
\label{sec:clip}
Evaluating geographical representativeness of text-to-image models through user studies is labor intensive, expensive and not easily reusable (for future models). 
It would be ideal to automatically 
quantify 
the geographical representativeness 
of unseen test images. 
In this section, we analyse the feasibility of such automatic evaluation (\textbf{RQ4} in Section \ref{sec:approach}). Particularly,
we explore automatically estimating the geographical representativeness using two different approaches: (i) using CLIP (a text-image alignment model) to  obtain the similarity between the country-specific textual prompt and the test image; and (b) using the similarity of the test image to already annotated images, i.e., via a $k$-nearest neighbor model. We elaborate these schemes below:

\subsection{CLIP-based Similarity}
One of the common techniques
used to automatically quantify biases in the text-to-image models is to 
use CLIP-based similarity as a proxy~\cite{radford2021learning}. For instance, CLIP similarity scores have been previously used to evaluate gender, racial, ethnic and cultural biases in text-to-image models~\cite{cho2022dall, bansal2022well, struppek2022biased}. Further, it has also been used to evaluate cross-lingual coverage of a concept in text-to-image models~\cite{saxonmultilingual}.
To assess if the CLIP model could be a useful tool for automatically estimating the geographical representativeness scores for a given country-noun pair, 
we use it to obtain the un-normalized similarity score between the image and a query of the form ``high definition image of a typical \texttt{[noun]} in \texttt{[country]}'', and compare it to the  geographical representativeness score assigned by participants from our study. 
We evaluate if we could reach the same findings (as in \S\ref{sec:results}) by using the CLIP similarity scores.

\noindent \textbf{Results.} 
~~Overall,
we find that the 
images generated through country-specific prompts 
have higher CLIP-based similarity scores 
than those generated by country-agnostic prompts (p-value $< 0.001$), 
for both \dalle and Stable Diffusion. 
Of all the cases
where \dalle images 
generated using country-specific prompts have a higher score
than images generated without country names, 
$98.7\%$ of the times the CLIP similarity scores are 
also higher. 
For the Stable Diffusion model,
the corresponding percentage is $96.4\%$.
These high-level findings are consistent with the user study. However, 
when we compare the scores of \dalle and Stable Diffusion models, 
CLIP-based similarity suggests that there is no statistically significant difference
in the geographical representativeness of images generated with country name,
which contradicts 
the results from the participants 
(they find images generated from \dalle with country-specific prompts 
to be more geographically representative than ones from Stable Diffusion).
Moreover, for images generated without the country name, the CLIP similarity scores are higher for Stable Diffusion than \dalle unlike the human ratings, for which there is no statistically significant difference.

Next, we 
study if we could obtain 
finer-grained 
findings 
similar to 
what we 
observe  
through a human study. 
For this, we first 
compute the Pearson's correlation coefficient, $\rho$,  
between country-wise 
geographical representativeness scores 
and CLIP similarity scores. 
We find no correlation 
across all nouns for images generated 
with country names ($\rho=0.01$), and weak 
correlation 
for images with country-agnostic prompts ($\rho=0.34$). 
Further, 
we  curate 
a
benchmark 
comprising pairs of images, 
and evaluate how often do human preferences 
(about which of the two images is more geographically representative)
match with the one selected through CLIP-based similarity.
We note that the 
agreement is merely $52.4\%$, 
where random chance agreement would be $50\%$.
These results 
indicate that 
the CLIP-based similarity 
is an inadequate  
proxy 
for the geographical representativeness.

 \subsection{Estimation using Nearest Neighbors} 
 We further explore the viability of  
 estimating the geographical representativeness 
 of a given test image (possibly generated by a future  text-to-image generation model) using the existing ratings 
 collected for images from \dalle and Stable Diffusion. 
 For a test image $X^T_n$ of a given noun $n$, 
 we define $\mathcal{X}^c_n$ as
 the set of images of $n$ annotated by participants of country $c$.
 Since a given image may be reflective 
 of surroundings in multiple countries, 
 we attempt to estimate the 
\textbf{GR} scores corresponding to all the studied countries. 
 For $X^T_n$, 
 we find its $k$ nearest neighbors by extracting the feature vectors 
of $X^T_n$ and the images in $\mathcal{X}^c_n$ from the vision model used by CLIP, 
and then computing the cosine similarities between the 
corresponding features. %
The predicted \textbf{GR} score of $X^T_n$ for country $c$
is the average of the human ratings corresponding to the obtained nearest neighbors.  
Specifically, 
we use the participant ratings of \dalle as the training data 
and those of Stable Diffusion for testing. Therefore, for noun $n$ and country $c$, $|\mathcal{X}^c_n| = 4$, as we have $4$ annotated images per noun for a given country, $2$ generated with country-specific prompts, the other $2$ generated without the country-specific prompts.
For example, 
to evaluate the \textbf{GR} score of an image of a house in India generated by Stable Diffusion, 
we find its $k$ nearest neighbors among the images that 
are generated through \dalle \emph{and} annotated by Indians. 
The estimated score is then compared 
to the true ratings of Indian participants.

\hfill \break
\noindent \textbf{Results.}~~Given that $|\mathcal{X}^c_n| = 4$, we set $k = 1$ for all our experiments. We find that the average correlation coefficient, the correlation between the human marked scores and the estimated scores is moderate ($\rho = 0.46$) over all the countries in the unspecified case, however, we find no correlation ($\rho = 0.01$) in the case of country-specific prompts. 
Further, the mean squared error (MSE) 
between the human and estimated scores is $1.39$ for images with country-agnostic prompts and $1.56$ for images with country-specific prompts. As a reference, we 
also check the MSE for a baseline value of 
$3.0$ for all the test images across all countries (as $3$ falls in the middle of $1$-$5$ scale).
For this reference, the MSE is $1.18$ for unspecified case and $0.83$ for country specific case---both these error values are lower than the 
corresponding values obtained using the estimates from the $k$ nearest neighbor model. 
These values point to the infeasibility of using 
this approach 
for automatically estimating the geographical representativeness, at least in the current form.
We believe that 
this 
is partly due to the fact that we 
only have a few 
annotated images in the training corpus to match with.
We also speculate that the image feature extractors (used for similarity computation)
may not extract features that differentiate 
images along the geographical lines.
We further present the MSE scores 
of the nearest neighbor method by varying the underlying pretrained feature extractor
in Table \ref{tab:feature-extractor}. 
We note that for both the country unspecified and the country specific cases, 
the MSE values for the predicted \textbf{GR} scores with respect 
to all the feature extractors are higher than that of values obtained using the baseline score of $3.0$.
This 
further underscores that 
automatically estimating geographical representativeness of images is challenging.

\begin{table}[]
\centering
\caption{Evaluating the estimated
 geographical representativeness using $k$-nearest neighbor approach. 
  We find the the Mean Squred Errors (MSE) for all the feature extractors are too high to be useful. 
  }
\label{tab:feature-extractor}
\begin{tabular}{@{}lll@{}}
\toprule
Approach & w/o country & w/ country \\ \midrule
Reference ($= 3.0$) & 1.18 & 0.83 \\ \hline  
Feature extractors:  & & \\ 
~~~~~VGG16 \cite{simonyan2014very}           & 1.55        & 1.52       \\
~~~~~ResNet18 \cite{he2016deep}        & 1.67        & 1.77       \\
~~~~~ResNet50 \cite{he2016deep}         & 2.04        & 1.62       \\
~~~~~ViT    \cite{dosovitskiy2020image}           & 1.81        & 1.51       \\
~~~~~CLIPVision \cite{radford2021learning}       & 1.38        & 1.56       \\ \bottomrule
\end{tabular}
\end{table}
Both the investigated approaches 
for estimating geographical 
representativeness 
turn out to be inadequate. 
We are able to reach similar high-level 
conclusions using CLIP-based similarity, but 
the similarity 
scores contradicted  
finer-grained findings. 
Overall, it is fundamentally 
challenging to automatically estimate 
the representativeness of images.

\section{Limitations \& Future Directions}
\label{sec:limitations}

\begin{figure}[h]
    \centering
    \includegraphics[width = \columnwidth]{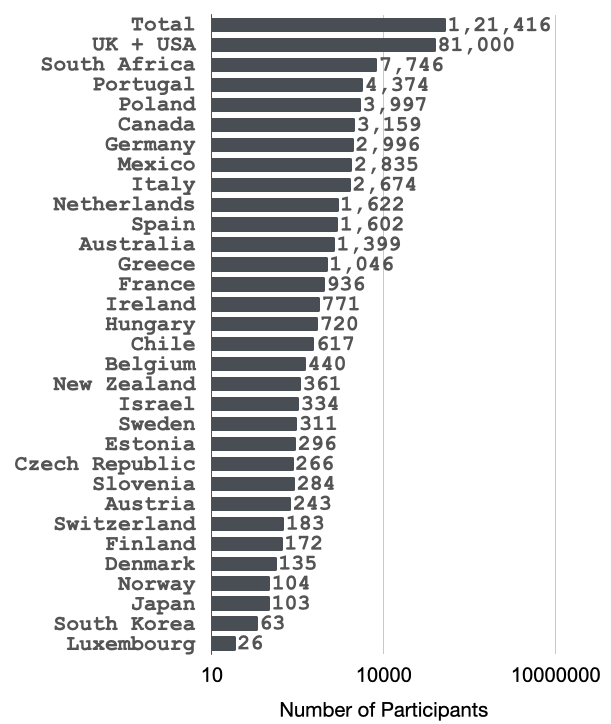}
    \caption{The number of participants available for research studies on Prolific are heavily skewed, and have few (or no) participants from many poor and developing nations. Such disparity is a serious challenge for inclusive model development and evaluation.}
    
    \label{fig:prolific}
\end{figure}

There are several important 
limitations 
of our work. 
Despite 
our efforts to reach out to participants from $88$ countries, 
we received sufficient responses 
from users only in
$27$ countries, and hence our study is limited to only $27$ countries.
We received \textbf{less than $5$
responses} from participants in 
Nepal (1), 
Bangladesh (2), 
Malaysia (2),
Turkey (5),
Singapore (2),
Argentina (1),
Kenya  (3),
Venezuela (1),
Pakistan (1),
Indonesia (2),
Nigeria (2),
Romania (2),
Colombia (3),
Namibia (1),
and \textbf{zero responses} 
from Laos, Armenia, Yemen, Thailand, Vietnam, Sri Lanka, Kazakhstan, Ukraine, Sierra Leone, Burkina Faso, Morocco, Senegal, Philippines, Egypt, Peru, Ethiopia, Mozambique, Kyrgyz Republic, Tanzania, Mali, Ecuador, Myanmar, Cambodia, Russia, Andorra, Finland, Tunisia, Gabon, Angola, Algeria, Libya, Botswana, and Seychelles.
As past surveys note, 
internet 
is not uniformly accessible 
across the globe~\cite{pew_research_internet_access, world_bank_internet_access}. The lack of 
access disproportionately
impacts 
marginalized and
poor nations,
which
further limits 
the voice residents 
of marginalized countries
have on the internet. 
Systems trained 
on the internet data run the risk 
of excluding such communities.
Perhaps due to internet access issues, 
crowdsourcing platforms 
have few (or no) 
participants from many developing countries, 
which further exacerbates
inclusive development
and evaluation of machine learning models (country-wise details can be found in Figure \ref{fig:prolific}).

Another weakness of our 
work is that we evaluate generated 
images for only $10$ common nouns.
As we evaluate two different models
with two different kinds of prompts
and use multiple images 
per noun, 
we end up with a survey comprising $80$ images 
per participant.
Including additional nouns 
would have resulted
in longer (or more) surveys 
and likely lower participation. 
However, we 
will open-source 
the code and required tools
for future work to reproduce and 
extend similar studies. 
An interesting future direction
is to examine 
techniques to aggregate 
images (for a given noun and a country) to speed and scale up the evaluation.

To improve the models, 
and the geographical representativeness of the generated images, we believe that more work is required to 
better document the sources of image-text pairs in the training data so as to understand the distributions of different objects and countries. 
Further, 
we need to 
collect and augment more data from the under-represented countries---there have been some past attempts at scraping more diverse image data~\cite{ramaswamy2023beyond}. Lastly, we call for improving  the participation from under-represented  countries in development and evaluation of machine learning models.

\section{Related Work}
\label{sec:related}

\noindent \textbf{Text-to-image Generation.} 
~Over the last few years,
models that convert any input text to
images have gained significant traction.
Initial text-to-image generation model used Generative Adversarial Networks \cite{zhang2017stackgan, reed2016generative, tao2020df, zhu2019dm} and Generative RNNs \cite{mansimov2015generating}. Recent advancements in transformers \cite{vaswani2017attention} and diffusion models~\cite{rombach2021highresolution}, and their application to text-to-image generation, has improved the quality of generated images. Autoregressive models encode the image as a grid of latent codes and train a multimodal transformer language model to generate the image tokens \cite{ramesh2021zero, cho2020x, yu2022scaling}. Another line of work employs diffusion models for image generation \cite{ruiz2022dreambooth, ramesh2022hierarchical, nichol2021glide, saharia2022photorealistic}. A different line of work fuses the diffusion models with autoregressive transformers \cite{gu2022vector}. For our study, we pick \dalle~\cite{ramesh2022hierarchical}, a diffusion based model released by OpenAI, and Stable Diffusion \cite{rombach2021highresolution}, an open-source latent text-to-image diffusion model, as there 
are increasing concerns that generated images from these models exhibit and amplify societal biases~\cite{bianchi2022easily, ramesh2022hierarchical, yu2022scaling}, since
they are trained on a large number of text-image pairs scrapped from the web and other sources.

\noindent \textbf{Societal Biases.}
~There is a growing body of work that 
critically analyzes the outputs of deep learning models in an attempt 
to discover and measure societal biases
for various downstream applications including
image classification~\cite{tong2020investigating, ramaswamy2021fair, 10.1145/3394171.3413772}, 
image captioning~\cite{zhao2021understanding, hendricks2018women}, 
language generation~\cite{sheng2021societal, sheng2019woman}, 
face recognition~\cite{buolamwini2018gender, conti2022mitigating}, 
image search~\cite{metaxa2021image}, and art creation~\cite{srinivasan2021biases}.
A recent study 
investigates
generated images from DALL·E v1 for gender and racial biases\cite{cho2022dall}, and another work 
examines the outputs of
Stable Diffusion models for stereotypes associated with gender, class, and ethnicity~\cite{bianchi2022easily}. The latter study~\cite{bianchi2022easily}
showcases several instances of dangerous biases exhibited by these models, and 
cautions against widespread adoption of such models.
Our study is similar in spirit to prior studies that aim to measure societal biases but 
analyzes---an oft-overlooked aspect of inclusive representation---geographical representation.

\section{Conclusion}
In this work, we investigated how well 
the images generated 
by two popular text-to-image 
models (\dalle and Stable Diffusion) 
reflect surroundings 
across the world. 
We conducted a user study involving $540$ participants 
from $27$ countries, wherein 
we asked participants 
the degree to which generated images of 
common nouns reflect their surroundings.
We found 
that when the input prompt
does not include any specific country name, 
users from $25$ out of $27$ countries felt 
that the generated images were 
less representative of the artifacts, with an average score of $2.39$. 
However, ratings increased to $3.49$ on an average 
when we included the country name in the text prompts.
These results also highlight how 
there is considerable room for models
to generate more geographically representative content.
When comparing \dalle with the Stable Diffusion model, we found that \dalle outperformed Stable Diffusion when using country specific inputs, but in other cases, these two models received similar scores. We 
also explored the feasibility of 
automating our study, and noted that the explored approaches were inadequate. Lastly, we highlighted key limitations and discussed ideas for future work to 
scale up the study and 
improve the geographical representativeness.

\section*{Acknowledgements}

We thank all the participants for their time and effort in scoring the images. We are grateful to Vinodkumar Prabhakaran, Sameer Singh, Preethi Seshadri and the members of the Vision and AI Lab, Indian Institute of Science, for the their valuable feedback.

{\small
\bibliographystyle{ieee_fullname}
\bibliography{main.bbl}
}
\onecolumn
\appendix
\section{Qualitative Analysis}
\begin{figure*}[h!]
    \nopagebreak
    \centering
    \includegraphics[width = \linewidth, height=0.85\textheight,trim={0.5cm, 0cm, 2cm, 0cm}, clip]{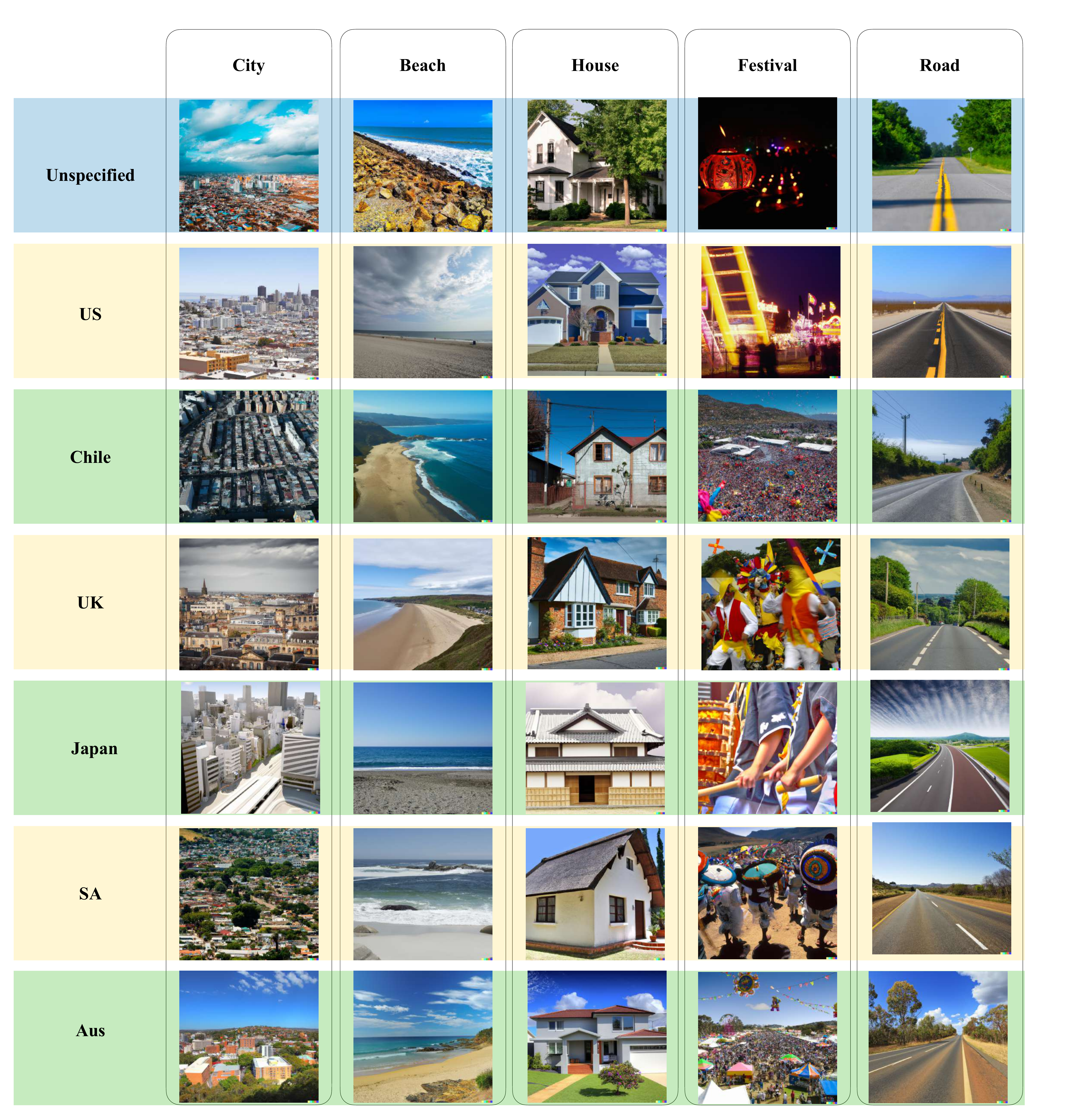}
    \caption{Qualitative examples of images of five common nouns (City, Beach, House, Festival and Road) generated by \dalle, for the default (unspecified) and the country specific prompts. From the shown examples, we note that the default generations often reflect artifacts from the US, whereas the representativeness is lower for Japan and Chile. The trend remains similar even after specifying the country name in the prompts, though the \textbf{GR} scores increase for all countries. (SA: South Africa, Aus: Australia)}
    \label{fig:d2-1}
\end{figure*}
\begin{figure*}[]
    \centering
    \includegraphics[width = \linewidth, height=0.9\textheight,trim={0cm, 0cm, 2.5cm, 0cm}, clip]{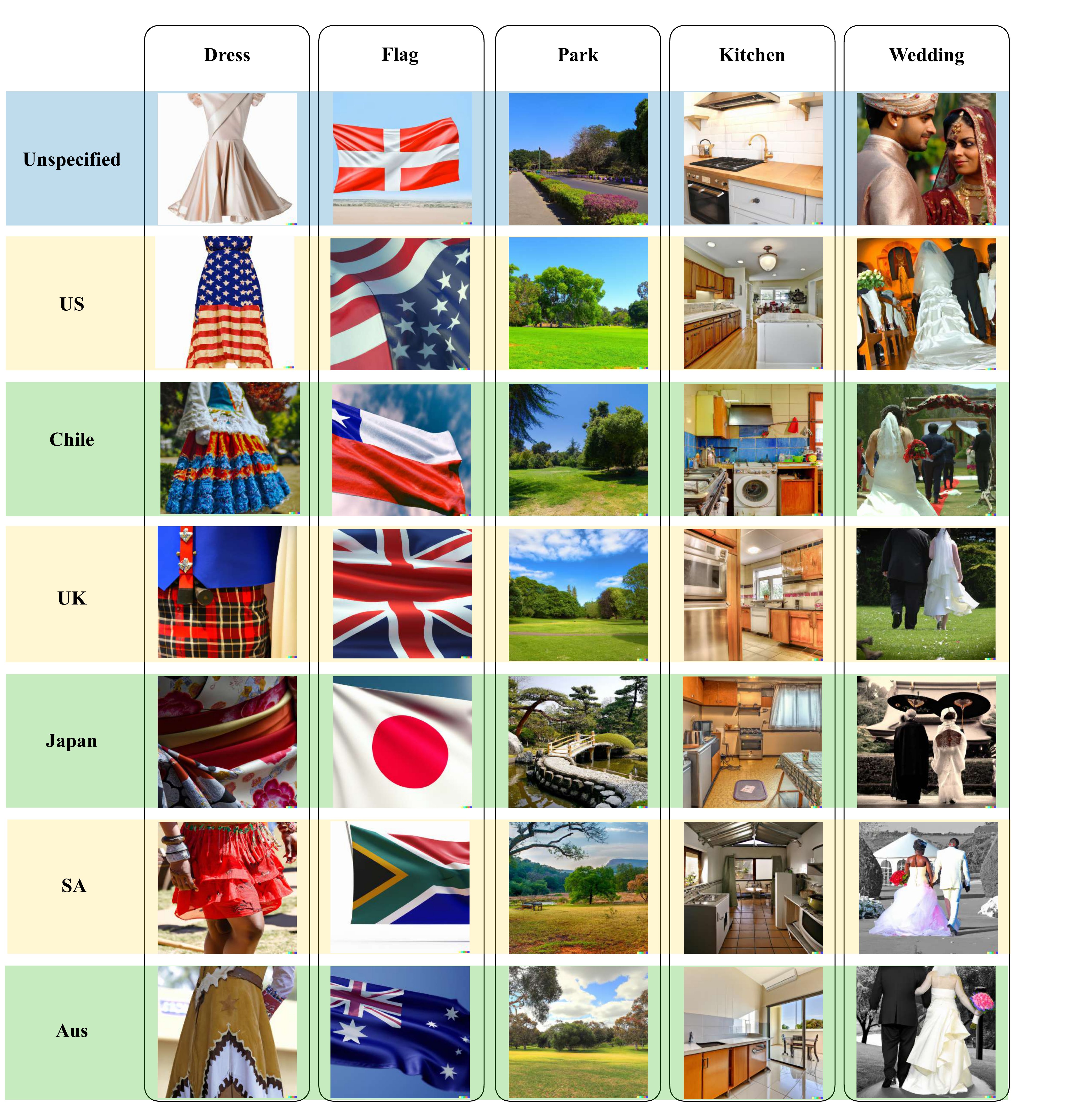}
    \caption{Qualitative examples of images of the five other common nouns (Dress, Flag, Park, Wedding and Kitchen) generated by \dalle for the default (unspecified) and the country specific prompts. From the shown examples, we note that the default generations often reflect artifacts from the US, whereas the representativeness is lower for Japan and Chile. The trend remains similar even after specifying the country name in the prompts, though the \textbf{GR} scores increase for all countries. (SA: South Africa, Aus: Australia)}
    \label{fig:d2-2}
\end{figure*}
\begin{figure*}[]
    \centering
    \includegraphics[width = \linewidth, height=0.9\textheight,trim={0cm, 0cm, 2cm, 0cm}, clip]{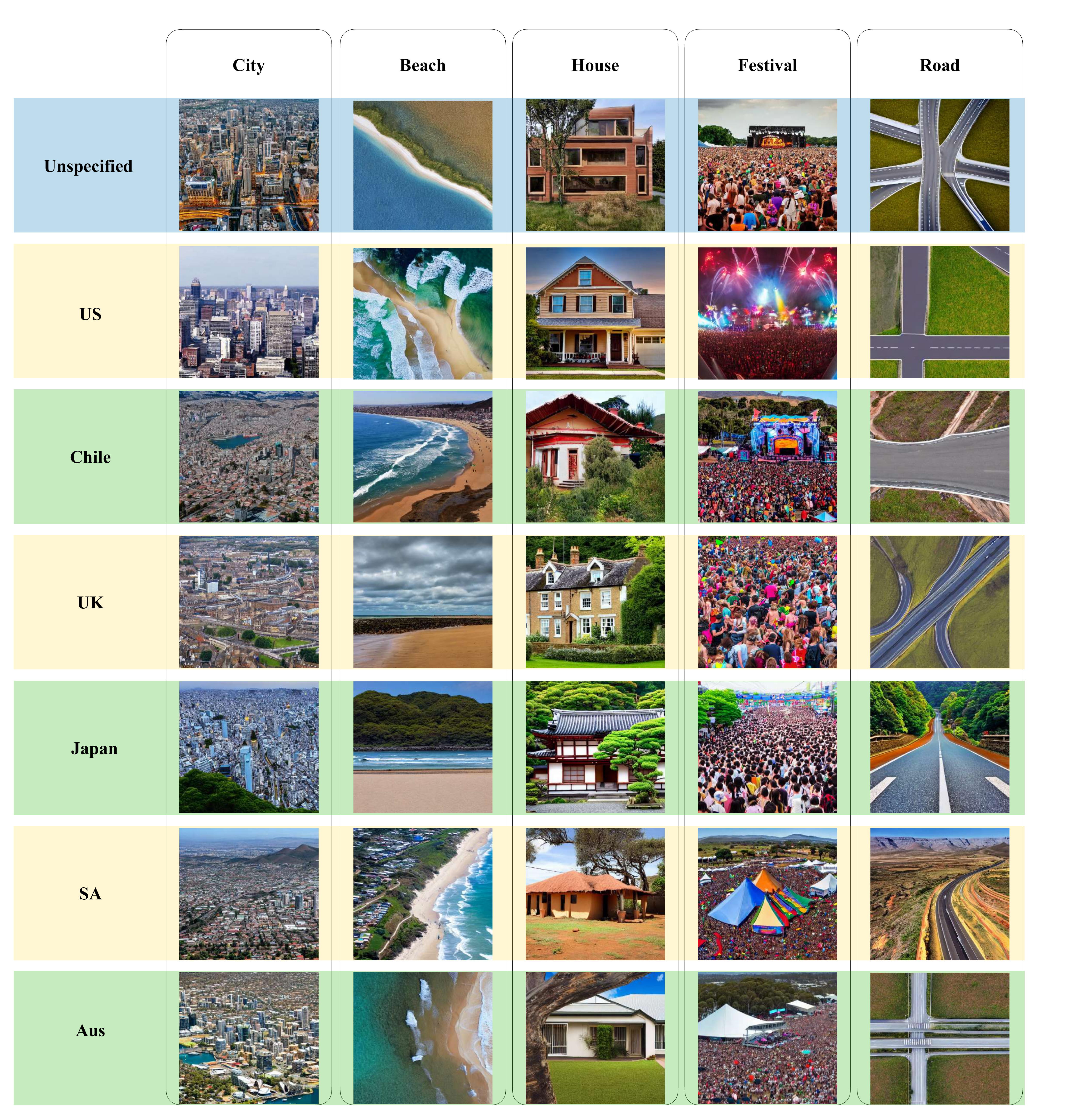}
    \caption{Qualitative examples of images of the five other common nouns (City, Beach, House, Festival, Road) generated by Stable Diffusion, for the default (unspecified) and the country specific prompts. Like the \dalle generated images, we note that the default generations often reflect artifacts from the US, whereas the representativeness is lower for Japan and Chile. The trend remains similar even after specifying the country name in the prompts, though the \textbf{GR} scores increase for all countries. (SA: South Africa, Aus: Australia)}
    \label{fig:SD-1}
\end{figure*}
\begin{figure*}[]
    \centering
    \includegraphics[width = \linewidth, height=0.9\textheight, trim={0cm, 0cm, 2.5cm, 0cm}, clip]{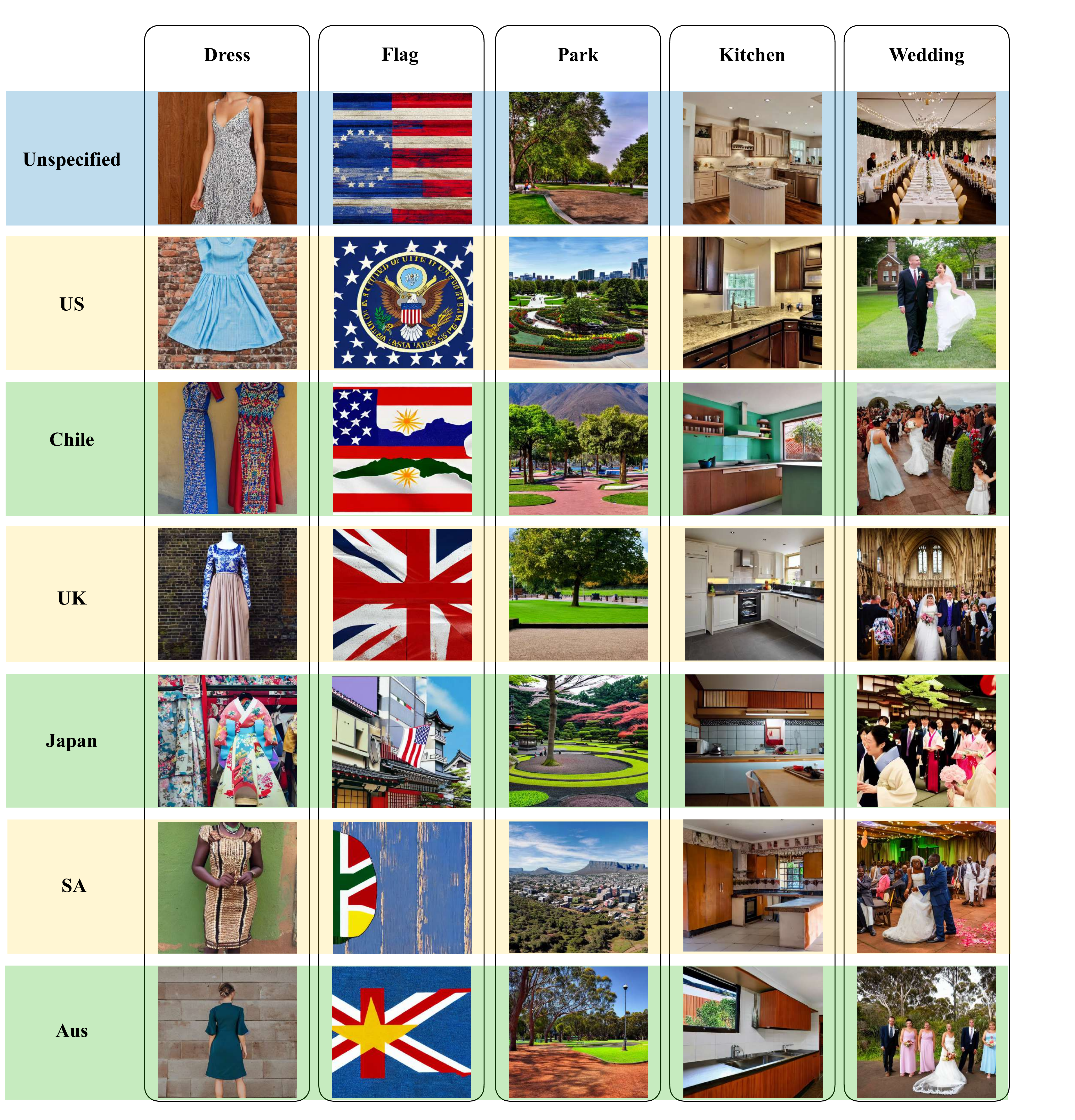}
    \caption{Qualitative examples of images of the five other common nouns (Dress, Flag, Park, Kitchen, Wedding) generated by Stable Diffusion, for the default (unspecified) and the country specific prompts. Once again, we note that the default generations often reflect artifacts from the US, whereas the representativeness is lower for Japan and Chile. The trend remains similar even after specifying the country name in the prompts, though the \textbf{GR} scores increase for all countries. (SA: South Africa, Aus: Australia)}
    \label{fig:SD-2}
\end{figure*}
\end{document}